%% file: arxiv.tex
\documentclass[10pt,twocolumn,letterpaper]{article}

 \usepackage{cvpr}              % To produce the CAMERA-READY version
\usepackage{algorithm}
\usepackage{algorithmic}

\input{preamble}
\input{math_cmd.tex}

\definecolor{cvprblue}{rgb}{0.21,0.49,0.74}
\usepackage[pagebackref,breaklinks,colorlinks,citecolor=cvprblue]{hyperref}

\def\confName{CVPR}
\def\confYear{2024}

\title{Single-Model and Any-Modality for Video Object Tracking}

\author{Zongwei Wu$^{1}$\quad Jilai Zheng$^{2}$ \quad  Xiangxuan Ren$^{2}$
    \quad Florin-Alexandru Vasluianu $^{1}$   \\  Chao Ma$^{2}$\thanks{Corresponding Author}  \quad Danda Pani Paudel$^{3}$ \quad Luc Van Gool$^{3,4}$ \quad Radu Timofte$^{1}$ \\
    \small $^{1}$ Computer Vision Lab, CAIDAS \& IFI, University of Wurzburg  \quad 
        $^2$ AI Institute, Shanghai Jiao Tong University \\ 
        \small  $^3$  INSAIT, Sofia University  \quad   \small  $^4$  CVL, ETH Zurich 
}

\begin{document}
\maketitle
\input{sec/0_abstract}

\input{sec/1_intro}
\input{sec/2_related}
\input{sec/3_methods}

\input{sec/4_exper}

{
    \small
    \bibliographystyle{ieeenat_fullname}
    \bibliography{main}
}

\end{document}

%% file: preamble.tex
\usepackage[dvipsnames]{xcolor}

\usepackage{times}
\usepackage{epsfig}
\usepackage{graphicx}
\usepackage{amsmath}

\usepackage{amssymb}

\usepackage{xcolor}
\usepackage{colortbl}
\usepackage{nicematrix}

\usepackage{textcomp}
\usepackage{stfloats}
\usepackage{url}
\usepackage{verbatim}
\usepackage{graphicx}
\usepackage{cite}
\usepackage{tikz}
\usepackage{comment}
\usepackage{color}
\usepackage{tabularx,lipsum}

\usepackage{wasysym}
\usepackage{graphics} % for pdf, bitmapped graphics files
\usepackage{epsfig} % for postscript graphics files
\usepackage{mathptmx} % assumes new font selection scheme installed
\usepackage{color} 
\usepackage{tabularx}
\usepackage{makecell}
\usepackage{multirow}
\usepackage{mwe,lipsum}
\usepackage{array,multirow}
\usepackage{float}
\usepackage{eucal}
\usepackage{comment}
\usepackage{pifont}

\usepackage{arydshln}
\usepackage[accsupp]{axessibility}

\definecolor{ao}{rgb}{0.0, 0.5, 0.0}

\newcommand{\tabincell}[2]{\begin{tabular}{@{}#1@{}}#2\end{tabular}}
\usepackage{multirow}
\usepackage{colortbl}
\definecolor{tabtitle}{gray}{.8}
\usepackage{bm}
\usepackage{pifont}

\usepackage{marvosym}

%% file: math_cmd.tex
\usepackage{amsmath,amsfonts,bm}

\def\eqref#1{equation~\ref{#1}}
\def\1{\bm{1}}

\def\vx{{\bm{x}}}

\DeclareMathAlphabet{\mathsfit}{\encodingdefault}{\sfdefault}{m}{sl}
\SetMathAlphabet{\mathsfit}{bold}{\encodingdefault}{\sfdefault}{bx}{n}

%% file: sec/0_abstract.tex
\begin{abstract}

In the realm of video object tracking, auxiliary modalities such as depth, thermal, or event data have emerged as valuable assets to complement the RGB trackers. In practice,  most existing RGB trackers learn a single set of parameters to use them across datasets and applications. However, a similar single-model unification for multi-modality tracking presents several challenges. These challenges stem from the inherent heterogeneity of inputs -- each with modality-specific representations, the scarcity of multi-modal datasets, and the absence of all the modalities at all times. In this work, we introduce Un-Track, a \underline{Un}ified Tracker of a single set of parameters for any modality. To handle any modality, our method learns their common latent space through low-rank factorization and reconstruction techniques. More importantly, we use only the RGB-X pairs to learn the common latent space. This unique shared representation seamlessly binds all modalities together, enabling effective unification and accommodating any missing modality, all within a single transformer-based architecture. Our Un-Track achieves \textbf{+8.1 absolute F-score} gain, on the DepthTrack dataset, by introducing only +2.14 (over 21.50) GFLOPs with +6.6M (over 93M) parameters, through a simple yet efficient prompting strategy. Extensive comparisons on five benchmark datasets with different modalities show that Un-Track surpasses both SOTA unified trackers and modality-specific counterparts, validating our effectiveness and practicality. The source code is publicly available at \url{https://github.com/Zongwei97/UnTrack}. 
\end{abstract}

%% file: sec/1_intro.tex
\section{Introduction}
\label{sec:intro}

\begin{figure}[t]
\centering
\includegraphics[width=\linewidth,keepaspectratio]{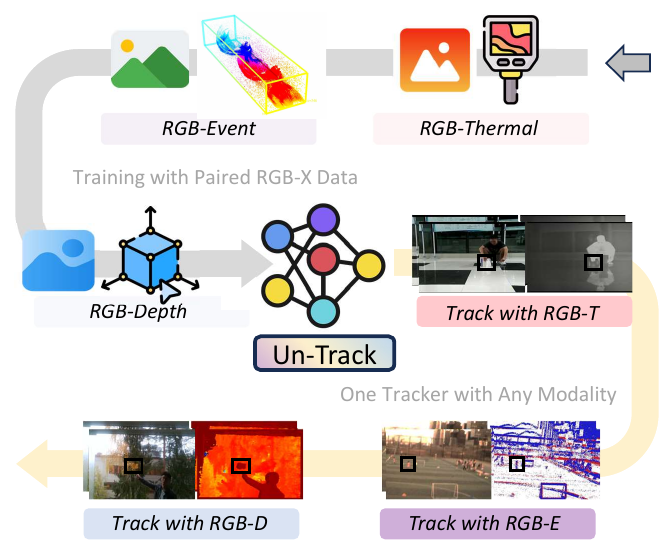}
\vspace{-8mm}
\caption{Un-Track is a unified tracker with a single parameter set that seamlessly integrates any modality (of RGB-X).}
\vspace{-4mm}
\label{fig:teaser}
\end{figure}

Video object tracking \cite{zhao2023representation,atom,SiamBAN,zhu2023tracking} is a fundamental task in computer vision with wide-ranging applications spanning from surveillance \cite{zheng2021robust} to augmented reality \cite{javed2022visual}, where accuracy and robustness are paramount. While traditional RGB trackers have shown promise in general settings, they often struggle with challenging scenarios such as occlusions \cite{stadler2021improving,lukezic2019cdtb}, low visibility \cite{zhu2023dcpt, zhang2022visible}, or fast-moving objects \cite{cioppa2022soccernet,tang2022coesot}. For more reliable tracking under such challenging conditions,  the integration of auxiliary modalities (X) like depth \cite{he2021fast,yang2022rgbd}, thermal \cite{zhang2022visible,zhao2021unified}, and event \cite{zhang2022spiking,zhu2023cross} have proven effective in multimodal tracking.

While the idea of fusing RGB with other modalities holds promise \cite{zhang2019robust,zhang2020multi}, the main challenge is the discrepancy in the representation of information across different modalities. Despite the success of previous fusion works \cite{fu2023distractor,xiao2022attribute} tailored for each specific scenario to improve RGB trackers, their reliance on modality-specific designs limits adaptability.  Recent initiatives \cite{protrack,zhu2023visual} towards a uniform architecture for various modalities show promise but necessitate modality-specific fine-tuning. This approach leads to multiple parameter sets, as shown in \cref{fig:teaser}(a), thereby compromising practicality in diverse real-world applications.

In this work, we aim to avoid such modality-specific fine-tuning to keep only one model-parameter set at all times. Another practical constraint arises from the differences in available auxiliary modalities across settings. Unifying modalities by a common representation can handle any modality at its disposal, addressing the mentioned problems by its virtue. However, two additional multifaceted challenges emerge from the scarcity of multimodal datasets and the absence of all paired data combinations. The former makes cross-modal mappings through a large-scale data prior unfeasible \cite{girdhar2023imagebind,lu2022unified,zhang2023all}, while the latter leads to missing modalities and renders joint learning using all possible combinations of paired data unfeasible \cite{zhang2023delivering,ma2021smil}.

To achieve such a unification, in this paper, we present \textbf{Un}-Track, denoting ``one" in French, which learns a cohesive embedding across diverse input modalities. Unlike previous approaches~\cite{zhang2023delivering,yin2021m2dgr,wisth2021unified,wan2023rpeflow}, Un-Track relies solely on RGB-X pairs for training, with X representing auxiliary modalities, without the need for all modalities to co-occur. Our objective is to discover a shared embedding seamlessly binding all auxiliary modalities (as depicted in \cref{fig:teaser}(b)). More specifically, we leverage the factorization prior, allowing reasoning about a common embedding directly from the low-rank latent space.  Factorization is a simple composition prior with the assumption that the global approximation can be constructed from a set of subset vectors. The factorization prior, successfully utilized in previous studies \cite{chen2020tensor,jhuo2012robust,Simon}, is employed in our work to reconstruct a shared embedding. This process transforms the heterogeneous modal representation into a uniform one, thereby facilitating the emergent cross-modal alignment.

Moreover, to harness the full potential of auxiliary inputs while maintaining efficiency, we leverage cross-modal features as prompts to enable RGB-X interaction. Different from previous works \cite{khattak2023maple, jia2022visual}, our goal is to enhance less reliable tokens, defined by a learnable score function, using multimodal cues. We approach this as a token recovery problem and leverage low-rank factorization to achieve the goal, which is first suggested in multimodal fusion, to the best of our knowledge. With its unified model architecture and prompting blocks, Un-Track is the first to offer support for cross-modality alignment under a single architecture with uniform parameters. In comparison to our RGB baseline with 21.50 GFLOPs and 92M parameters, Un-Track introduces only +2.14 GFLOPs with +6.6M parameters, resulting in a significant +8.1 absolute F-score gain demonstrated on the DepthTrack dataset. Extensive comparisons across five datasets with different modalities validate Un-Track's superiority over specialized SOTA models, surpassing both unified trackers and modality-specific fine-tuned counterparts by a substantial margin.

%% file: sec/2_related.tex
\section{Related Works}
\label{sec:related}

\noindent \textbf{Multimodal tracking:} Video object tracking \cite{chen2022high,yao2020video} aims to detect objects in a video sequence based on their initial positions. Early approaches treated tracking as a per-frame target-matching problem, with Siamese networks \cite{siamatten,siameserpn,siamrpnplusplus,SiamRCNN,siamfc++,Deeper-wider-SiamRPN} being a notable example. More recently, transformer-based methods \cite{yan2021learning,transt,chen2021transformer,trdimp,chen2022backbone,cui2022mixformer} have gained popularity for feature extraction and per-frame correlation in tracking. Large-scale training datasets \cite{trackingnet,lasot,got10k,Youtube} have empowered RGB trackers to uniformly apply parameters across various applications. While RGB trackers deliver promising results, challenges such as occlusion, low illumination, and fast-moving scenes have led to the exploration of additional modalities. Several works have investigated the role of depth \cite{zhu2023rgbd1k,yang2023resource}, thermal \cite{liu2020learning,long2019multi}, and event modalities \cite{tang2022revisiting,visevent} in enhancing tracking performance. Specifically, depth cues \cite{yan2021depth,gao2022learning} contribute to handling objects with different camera distances; thermal cameras \cite{wang2022mfgnet,xiao2022attribute} address challenges such as low illumination; event cameras improve the temporal awareness  \cite{zhu2022learning,zhang2021object,wang2022exploiting}.

Despite the plausible achievements, many rely on modality-specific blocks designed for individual modalities \cite{zhao2021adaptive,zhu2021robust}, limiting their adaptability. Recent efforts have focused on achieving architectural unification \cite{protrack,zhu2023visual}, yet they still necessitate modality-specific fine-tuning, resulting in distinct parameter sets for different modalities. The ideal scenario would involve a large-scale dataset encompassing all possible modal combinations, but current tracking datasets predominantly feature a single modality — depth \cite{rgbd1k,depthtrack}, thermal \cite{rgbt234,lasher}, or event \cite{visevent} — posing challenges for a unified model with a single parameter set.

\noindent \textbf{Learning with Missing modalities:} Recent research has addressed real-world scenarios where models must cope with missing modalities \cite{ma2021smil,qiu2023modal}. One common strategy involves estimating missing values by learning joint multimodal representations \cite{lee2023multimodal,zeng2022tag}, feasible when complete samples are available during training. However, tracking datasets typically exhibit only one modality at a time, complicating the learning of such joint representations. Other works \cite{lu2022unified, girdhar2023imagebind} implicitly learn cross-modal alignment end-to-end using large-scale datasets and deep networks, demanding substantial computational resources. Extending such approaches to tracking is challenging due to limited downstream datasets and real-world applicability constraints \cite{yan2021lighttrack,kang2023exploring}. In contrast to existing methods, our approach investigates cross-modal relationships by leveraging edge priors to learn a joint representation that unifies all modalities. Our method does not require the simultaneous occurrence of all modalities, offering a unique perspective on dealing with diverse and individual modalities.

%% file: sec/3_methods.tex
\section{Methods}

\begin{figure*}[t]
\centering
\includegraphics[width=.9\linewidth,keepaspectratio]{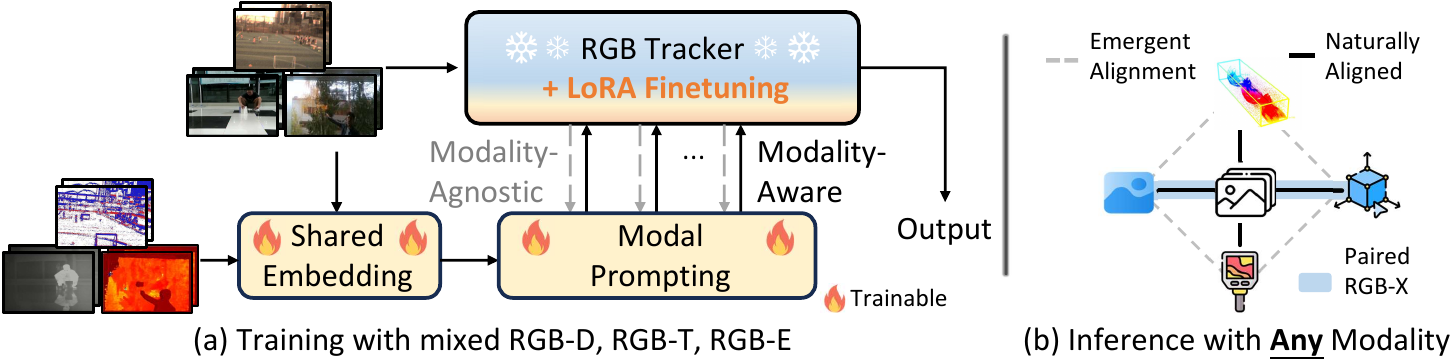}
\vspace{-3mm}
\caption{Our \textbf{proposed framework}, termed Un-Track, is composed of a shared embedding, a modal prompting, and a LoRA-finetuned pretrained RGB tracker. The shared embedding learns a joint representation that unifies all modalities (\cref{sec:recons}). The modal prompting block enhances feature modeling with modal awareness at each scale (\cref{sec:prompt}). To track the target object, we finetune a pretrained foundation model \cite{ostrack} using the LoRA technique (\cref{sec:lora}). Our model achieves a unified model applicable across different modalities under a single parameter set.  During inference, Un-Track seamlessly integrates any image-paired data, thanks to the emergent alignment.}
\vspace{-3mm}

\label{fig:model}
\end{figure*}

\subsection{Overall Framework}
In this paper, our primary focus is on multimodal tracking, with the constraint that only one modality is available at a time, as shown in \cref{fig:model}. We define our multimodal tracking dataset as $M= \{M^D, M^T, M^E\}$, where $M^X$ represents the subset dataset with only a single modality $X$ available. Conventional methods tackling missing modalities often use dummy inputs for the absent pixels \cite{lee2023multimodal,wang2023multi}, simulating complete datasets with all possible combinations of paired datasets. In contrast, our method transforms any auxiliary input into a shared embedding, seamlessly binding all modalities together and creating a complete and paired representation with the master RGB feature.

To mitigate overfitting on sparse downstream multimodal datasets, we adopt a transformer-based RGB tracker with frozen parameters and fine-tune it for multimodal tracking. Leveraging a lightweight outer prompting method, we identify uncertain tokens at each scale and enhance them with cross-modal awareness. Simultaneously, an inner fine-tuning process is implemented using the LoRA technique \cite{hu2021lora}.

During training, our model learns the shared embedding from samples in the mixed dataset $M$, effectively binding all modalities together. As for inference, our model can accommodate any modal input $X$, thanks to the emergent alignment. Our trainable parameters only include cross-modal binding, outer prompting parameters, and inner LoRA parameters, ensuring a training-friendly pipeline that can be efficiently employed end-to-end on a single 24G GPU.

\subsection{Shared Embedding}
\label{sec:recons}

\noindent \textbf{Explicit Edge Awareness:} We observe that, as illustrated in \cref{fig:model}(a), depth data introduce 3D distance information, effectively delineating objects with varying granularity and enhancing the sharpness of 3D boundaries; thermal images generate a scene heat map, highlighting objects based on their temperatures and providing clearer contours; event data capture intensity changes, particularly around an object's outbound region. Notably, a consistent feature emerges across these modalities: the representation of the ``true" objects' shape, often manifested through edges.

\begin{figure}[t]
\centering
\includegraphics[width=\linewidth,keepaspectratio]{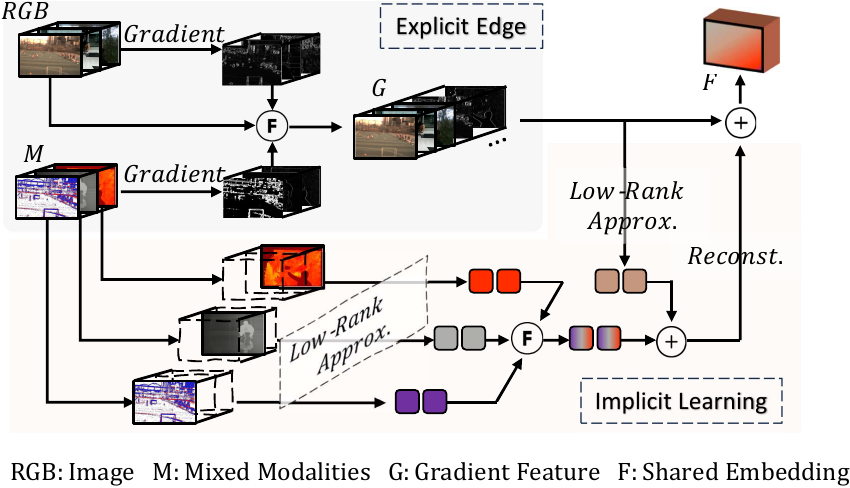}
\vspace{-6mm}
\caption{\textbf{Shared Embedding}. We derive a joint representation through low-rank factorization and reconstruction. Such an implicit learning is additionally integrated with explicit edge awareness to enhance the embedding.}
\label{fig:shared}
\vspace{-4mm}
\end{figure}

Motivated by this observation, our objective is to harness edge embedding to unify all modalities. To achieve this, as shown in \cref{fig:shared}, we generate gradient maps from auxiliary modalities by computing differences between neighboring pixels along both the x- and y-axes. Simultaneously, without compromising texture edge, we also generate RGB gradient maps. Subsequently, all gradient maps are integrated with the visual feature, forming the gradient feature $G$.

\noindent \textbf{Implicit Low-Rank Reconstruction:} While edges present a shared feature across different modalities, exclusively transforming all modalities into edges may risk overlooking modality-specific clues. Therefore, we introduce an implicit learning pipeline to complement this by discovering the shared embedding, guided by the previously generated explicit edge awareness. This combined approach allows for a more effective identification of the shared embedding, leveraging both data-driven learning and edge priors.

In technical terms, we redefine the challenge of learning the shared embedding as a quest for the shared low-rank approximation. Both objectives share the essence of distilling common features across all modalities. However, direct estimation of the low-rank approximation becomes impractical due to the distinct data domains and modal representations. In response, we propose a pragmatic strategy: decomposing the shared low-rank vector into the low-rank of each subset component. This alternative, more manageable and feasible within a single domain, lays the groundwork for approximating the global shared low-rank from these individual low-rank components.

The overall algorithm can be found in \cref{algo}. Specifically, let $M$ be the input feature with mixed auxiliary modalities, decomposed into subset features $D, T, E$ representing depth, thermal, and event samples from subset datasets ${M^D, M^T, M^E}$. Their respective $k_{th}$ low-rank matrices $D_k, T_k, E_k$, are approximated by:
\begin{equation}
    D_k = \sigma_d(D), \quad T_k = \sigma_t(T), \quad E_k = \sigma_e(E),
\end{equation}
where $\sigma_x$ is the modality-specific learning through a simple MLP projecting features from input channel $c$ to the low-rank space $k$ ($k < c$). Simultaneously, we compute the low-rank matrix $G_k$ from the gradient feature $G$.

\begin{algorithm}[t]
{
\small
\caption{Implicit Shared Embedding}
\label{algo}
\begin{algorithmic}
\label{alg:edgeLoraForward}
\STATE \textbf{Input:} $M$ - Mix Modal Feature, $G$ - Explicit Gradient Binding.
\STATE \textbf{Output:} $F$ - Reconstructed feature

\STATE 1. \textbf{Initialize:} Separate $M$ into subsets, each representing a modality-specific feature: $D, T, E$.

\STATE 2. \textbf{In-Domain Approximation:} Train individual approximators $\sigma_x$ for each modality $x$ to derive domain-specific low-rank matrices $D_k, T_k, E_k$. Idem for explicit low-rank gradient $G_k$.

\STATE 3. \textbf{Fuse and Guide:} Merge $D_k, T_k, E_k$ through a fusion function $\varphi_1$. Incorporate the explicit gradient $G_k$ using addition, after projecting it into a compatible latent space through $\varphi_2$.

\STATE 4. \textbf{Reconstruct:} Train a reconstruction function $\Phi_R$ to construct the shared embedding $F$, considering the explicit guidance $G$ in the original feature space.

\end{algorithmic}
}
\end{algorithm}

The global shared low-rank matrix $M_k$ is then approximated by fusing the subset low-rank matrices $D_k, T_k, E_k$, along with the gradient guidance. Technically, we concatenate $D_k, T_k, E_k$ and learn the joint low-rank approximation, incorporating it with the gradient guidance. This pipeline can be expressed as follows: 
\begin{equation}
M_k = \varphi_{R_1}([D_k, T_k, E_k]) + \varphi_{R_2}(G_k),
\end{equation}
where $[.]$ is the channel concatenation and $\varphi_{R_i}$ are the MLP projections to the low-rank latent space. Finally, we reconstruct the shared embedding $F$ through:
\begin{equation}
    F = \Phi_R(M_k) + G,
\end{equation}
where $\Phi_R$ is another MLP that projects the jointly-learned low-rank matrix back to the departing feature space. Our ablation studies validate that this subset low-rank approximation and regrouping efficiently unify all input modalities, despite their heterogeneous representations.

\begin{figure}[t]
\centering
\includegraphics[width=\linewidth,keepaspectratio]{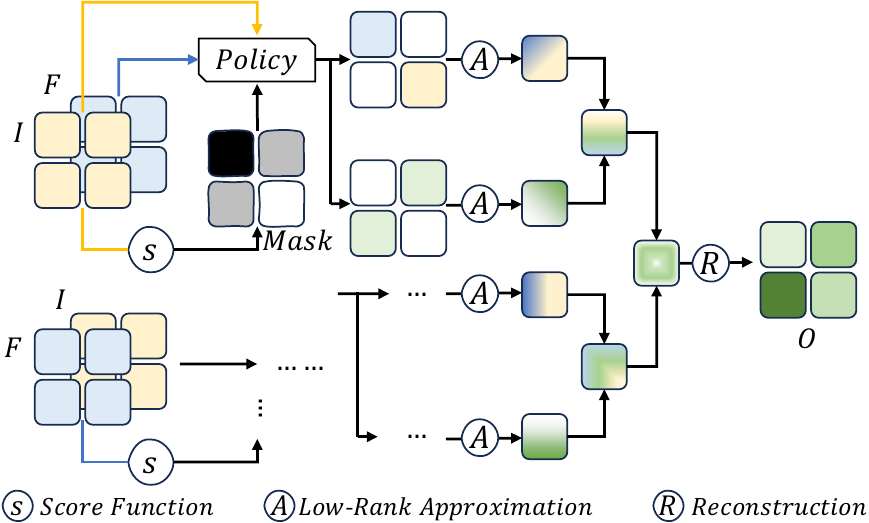}
\vspace{-5mm}
\caption{\textbf{Modal Prompting.} For the visual feature 
$I$, we employ a score function to categorize tokens into negative, uncertain, and positive segments. Using a token exchange policy, we discard negative tokens, enhance uncertain ones with corresponding tokens from $F$, and retain positive ones. Then, we transform the feature fusion task into a token recovery problem, addressed by low-rank factorization. Similarly, we extract the most informative low-rank matrix from $F$ to fuse and reconstruct the shared output.}
\label{fig:prompt}
\vspace{-2mm}
\end{figure}

\subsection{Outer Modal Prompting}
\label{sec:prompt}

RGB-tracker may fail to perform accurately in corner cases where auxiliary clues can contribute. Drawing inspiration from the success of adapting large pre-trained models to specific downstream tasks \cite{hu2021lora,jia2022visual}, we introduce a modal prompting method devised to enhance RGB token $I$ with modality-awareness $F$, as shown in \cref{fig:prompt}.

Specifically, our approach employs a shrinkage token fusion strategy. Taking $I$ as an example, we categorize tokens into three groups—negative, uncertain, and positive—based on a dynamic scoring function $s$. These regions are defined with mask form, expressed as $m_n, m_u, m_p = s(I)$. To harness multimodal clues effectively, we replace negative tokens with those from the other modality, omit uncertain ones with dummy values, and retain the positive tokens. Subsequently, these modified tokens undergo projection into the low-rank space using the approximation function $\sigma_c$. Our objective is to enhance robustness by completing uncertain tokens with information from reliable neighboring tokens. Mathematically, we obtain the first low-rank matrix $I_{l_1}$ by:
\begin{equation}
 I_{l_1} = \sigma_c(m_n \cdot F + m_p \cdot I).
\end{equation}

Next, we target the uncertain tokens by merging them with the paired tokens from the other modality and approximate the low-rank matrix similarly. Here, we aim to throw out the possible noise, resulting in a matrix that is more informative than the original. Let $\sigma_n$ be another approximation function, we obtain the second low-rank matrix $I_{l_n}$ by:
\begin{equation}
 I_{l_2} = \sigma_n(m_u \cdot F + m_u \cdot I).
\end{equation}

Then, we fuse these two low-rank matrices in low-rank space, forming the shared low-rank matrix $I_l$ for input $I$. In such a manner, we improve the image token modeling by fully benefiting from the auxiliary clues. Mathematically, the whole process can be formulated as:
\begin{equation}
I_l = \varphi_{P}([ I_{l_1}, I_{l_2}]),
\end{equation}
where $\varphi_{P}$ is learnable fusion. Similarly, from the input $F$, we follow the same process to obtain the low-rank matrix $F_l$. For the cross-modal fusion, we add these two low-rank matrices and then project back to the original space. We can obtain the fused output $O$ by:
\begin{equation}
    O = \Phi_P(I_l + F_l),
\end{equation}
where $\Phi_P$ is another learnable MLP. 

Our method can be regarded as a mixer of token exchange (for negative tokens) and token fusion (for uncertain tokens), while retaining the most informative modality-specific clues (for positive tokens). As the majority of fusion operations occur in the low-rank feature space, our progressive cross-modal shrinkage avoids imposing a significant additional computational burden, while being able to excavate and accumulate the rich cues from each modality for effective modal prompting.

\subsection{Inner Finetuning}
\label{sec:lora}

In addition to the outer modal prompting, we incorporate the LoRA technique \cite{hu2021lora} for more efficient finetuning. For each transformer attention module with the weight matrix $W_0\in \mathbb{R}^{d\times k}$, we introduce two learnable matrices: $B\in \mathbb{R}^{d\times r}$ and  $A\in \mathbb{R}^{r\times k}$. This leads to the replacement of the frozen attention mechanism $h = W_0 \displaystyle \vx$ with the new LoRA attention:

\begin{equation}
    h = W_0 {\displaystyle \vx} + BA{\displaystyle \vx}.
\end{equation}

To train our network, we adopt the same loss functions as our baseline tracker \cite{ostrack} for end-to-end learning.

%% file: sec/4_exper.tex
\begin{table*}[t]
	\small
	\centering
	\renewcommand\arraystretch{1.15} 
	\setlength{\tabcolsep}{3mm}{
		\resizebox{\linewidth}{!}{
			\begin{tabular}{c|ccccccc|ccccccc}
                        \toprule
				\small
   & \multicolumn{7}{c|}{Depth-Specific Parameters} & \multicolumn{7}{c}{Uni-model with a Single Set of Parameters}
    \\
                        \midrule
				&\tabincell{c}{ATCAIS\\~\cite{vot20}}&\tabincell{c}{DDiMP\\~\cite{vot20}}&\tabincell{c}{DeT\\~\cite{depthtrack}}&\tabincell{c}{SPT\\~\cite{rgbd1k}}&\tabincell{c}{ProTrack\\~\cite{protrack}}&\tabincell{c}{ViPT\\ \cite{zhu2023visual}} &\tabincell{c}{\textbf{Un-Track}\\\textbf{(ours)}} &\tabincell{c}{Stark\\ \cite{stark}} &\tabincell{c}{AiATrack\\ \cite{gao2022aiatrack}} &\tabincell{c}{OSTrack\\ \cite{ostrack}} &\tabincell{c}{UniNext\\ \cite{yan2023universal}}  &\tabincell{c}{SeqTrack \\ \cite{chen2023seqtrack}}  &\tabincell{c}{ViPT\\ \cite{zhu2023visual}} &\tabincell{c}{\textbf{Un-Track}\\\textbf{(ours)}}\\
                        \midrule
				F-score($\uparrow$)&0.476&0.485&0.532&0.538&0.578 &\textbf{\textcolor[rgb]{0,0,1}{0.594}} &\textbf{\textcolor[rgb]{1,0,0}{0.612}} &0.397&0.515&0.569&0.422&0.590&0.561&\textbf{\textcolor[rgb]{0,1,0}{0.610}} \\
				Re($\uparrow$)&0.455&0.469&0.506&0.549&0.573&\textbf{\textcolor[rgb]{0,0,1}{0.596}} &\textbf{\textcolor[rgb]{1,0,0}{0.610}} &0.406&0.526&0.582&0.432&0.600&0.562&\textbf{\textcolor[rgb]{0,1,0}{0.610}}\\
				Pr($\uparrow$)&0.500&0.503&0.560&0.527&0.583&\textbf{\textcolor[rgb]{0,0,1}{0.592}} & \textbf{\textcolor[rgb]{1,0,0}{0.613}} &0.388&0.505&0.557&0.413&0.580&0.560&\textbf{\textcolor[rgb]{0,1,0}{0.610}}\\
                        \bottomrule
			\end{tabular}
	}}\\
	\vspace{-3mm}
	\caption{
		\small
		Overall performance on DepthTrack test set~\cite{depthtrack}. \textcolor[rgb]{1,0,0}{Red}/\textcolor[rgb]{0,1,0}{Green}/\textcolor[rgb]{0,0,1}{Blue} stands for the best/second/third performance. Our depth-specific model sets the SOTA record, while our uni-model with single parameters set also outperforms previous depth-specific SOTA.}
	\label{tab-depthtrack}
	\vspace{-3mm}
\end{table*}

\section{Experiments}

\subsection{Training Data} 
%  our training is conducted solely on RGB+X paired data, comprising RGB-D, RGB-T, and RGB-E combinations. Different from previous approaches that fine-tune separate sets of parameters for each modality, Un-Track is trained on all modalities with a unified set of parameters. 
% Note that in our experimental setting, we have only one RGB+X (i.e. RGB+depth, \textbf{\textit{or}} RGB+thermal, \textbf{\textit{or}} RGB+event) at a time.

In the absence of a comprehensive multi-modal tracking dataset encompassing all possible combinations (RGB-D-T-E), we have only one RGB+X (i.e. RGB+depth, \textbf{\textit{or}} RGB+thermal, \textbf{\textit{or}} RGB+event) at a time. The conventional modality-specific adapted settings are referred as ``X-Specific", whereas the main target of this paper with 
 one model trained on all modality pairs is called ``Uni-model" (or a single set of parameters).   The training and evaluation settings are summarized as follows:
\vspace{-3mm}
\begin{table}[h]
\begin{tabular}{  m{2.5cm}|| m{2cm} | m{2cm} }
\hline
Model & Trained on & Evaluated on\\
\hline
\hline
X-Specific & \text{RGB}$_i$+$X_i$ & \text{RGB}$_i$+$X_i$\\
\hline
Uni-model & $\bigcup_i$\text{RGB}$_i$+$X_i$ & \text{RGB}$_i$+$X_i$\\
\hline
\end{tabular}
\vspace{-3mm}
\end{table}

Our RGB-D samples are sourced from DepthTrack~\cite{depthtrack}, a pioneering RGB-D tracking benchmark with 150 training long-term sequences. RGB-T samples are extracted from the extensive LasHeR~\cite{lasher} dataset, featuring 979 diverse training sequences. RGB-E samples are obtained from VisEvent~\cite{visevent}, which boasts 500 real-world sequences. Each D/T/E input is transformed into an RGB-like form.

\subsection{Within distribution Evaluation} 

Given that DepthTrack \cite{depthtrack}, Lasher \cite{lasher}, and VisEvent \cite{visevent} provide domain-specific testing sequences, our initial evaluation focuses on these within distribution sequences. For each dataset, we adhere to the metrics specified in the original papers and prior standards \cite{protrack,zhu2023visual} for evaluation.

\noindent \textbf{Comparison on DepthTrack \cite{depthtrack}}: For evaluation, we use metrics such as precision (Pr) and recall (Re), as well as the F-score, which are the primary metrics. As shown in \cref{tab-depthtrack} When exclusively trained and fine-tuned on DepthTrack, Un-Track achieves a +2.1\% absolute precision improvement over the current depth-specific SOTA ViPT \cite{zhu2023visual}. Notably, even when trained with mixed data using a single parameter set, Un-Track outperforms the depth-specific ViPT. Furthermore, when the current ViPT is jointly trained on all datasets with a single set of parameters, its performance significantly deteriorates.

Additionally, we observe that trackers excelling in other tracking datasets \cite{lasot,got10k,trackingnet} might struggle in RGB-D downstream settings. UniNext \cite{yan2023universal}, a leading tracker trained on various large-scale tracking datasets and related image/ video tasks, exhibits poor performance.  In contrast, our model achieves cross-modal unification within a single set of parameters, surpassing all depth-specific counterparts and performing closely to our specialized version. This underscores the efficacy of our shared embedding in achieving global alignment across diverse modalities.

\begin{table*}[t]
	\small
	\centering
	\renewcommand\arraystretch{1.15} 
	\resizebox{\linewidth}{!}{
		\begin{tabular}{c|cccccccc|cccccc}
                        \toprule
			\small

          & \multicolumn{8}{c|}{Thermal-Specific Parameters} & \multicolumn{6}{c}{Uni-model with a Single Set of Parameters}
    \\
                        \midrule

			&\tabincell{c}{SGT\\~\cite{sgt}}  &\tabincell{c}{FANet\\~\cite{fanet}}  &\tabincell{c}{mfDiMP\\~\cite{mfdimp}} &\tabincell{c}{DAFNet\\~\cite{dafnet}} &\tabincell{c}{MaCNet\\~\cite{macnet}}  &\tabincell{c}{ProTrack\\~\cite{protrack}} &\tabincell{c}{ViPT \\ \cite{zhu2023visual}}  &\tabincell{c}{\textbf{Un-Track}\\\textbf{(ours)}}  &\tabincell{c}{Stark\\ \cite{stark}} &\tabincell{c}{AiATrack\\ \cite{gao2022aiatrack}} &\tabincell{c}{OSTrack\\ \cite{ostrack}}  &\tabincell{c}{SeqTrack \\ \cite{chen2023seqtrack}} &\tabincell{c}{ViPT\\ \cite{zhu2023visual}} &\tabincell{c}{\textbf{Un-Track}\\\textbf{(ours)}}\\ 
                        \midrule
			PR($\uparrow$) &0.327 & 0.441 &0.447 &0.448 &0.482  &0.538 &\textcolor{green}{\textbf{0.651}} & \textcolor{red}{\textbf{0.667}}& 0.418 &0.463 & 0.530 &0.582 & 0.608 & \textcolor{blue}{\textbf{0.646}}\\
			SR($\uparrow$) &0.232 &0.309 &0.343 &0.311 &0.350 & 0.420 &\textcolor{green}{\textbf{0.525}} & \textcolor{red}{\textbf{0.536}} &0.333 & 0.365 &0.422 &0.441 & 0.490 & \textcolor{blue}{\textbf{0.513}} \\
                        \bottomrule
		\end{tabular}
	}
	\vspace{-3mm}
	\caption{
		\small
		Overall performance on the Lasher thermal testing set \cite{lasher}. Our thermal-specific model sets a new SOTA record. Our uni-model variant competes strongly with the previous SOTA thermal-specific models and significantly surpasses its unimodel version.
}
	\label{tab-lasher}
	\vspace{-1mm}
\end{table*}

\begin{table*}[t]
	\small
	\centering
	\renewcommand\arraystretch{1.15} 
 	\setlength{\tabcolsep}{1.4mm}
	\resizebox{\linewidth}{!}{
		\begin{tabular}{c|cccccccc|cccccc}
                        \toprule
			\small

          & \multicolumn{8}{c|}{Event-Specific Parameters} & \multicolumn{6}{c}{Uni-model with a Single Set of Parameters}
    \\
                        \midrule

			  &\tabincell{c}{SiamCar\\~\cite{stark}}&\tabincell{c}{Stark\_E\\~\cite{stark}}  &\tabincell{c}{VITAL\_E\\~\cite{song2018vital}}  &\tabincell{c}{PrDiMP\_E\\~\cite{danelljan2020probabilistic}} &\tabincell{c}{TransT\_E\\~\cite{chen2021transformer}}  &\tabincell{c}{ProTrack\\~\cite{protrack}} &\tabincell{c}{ViPT \\ \cite{zhu2023visual}}  &\tabincell{c}{\textbf{Un-Track}\\\textbf{(ours)}}  &\tabincell{c}{Stark\\ \cite{stark}} &\tabincell{c}{AiATrack\\ \cite{gao2022aiatrack}} &\tabincell{c}{OSTrack\\ \cite{ostrack}}  &\tabincell{c}{SeqTrack \\ \cite{chen2023seqtrack}} &\tabincell{c}{ViPT\\ \cite{zhu2023visual}} &\tabincell{c}{\textbf{Un-Track}\\\textbf{(ours)}}\\ 
                        \midrule
			Precision($\uparrow$) &0.599 &0.612 & 0.649  &0.644 &0.650  &0.632 &\textcolor{green}{\textbf{0.758}} & \textcolor{red}{\textbf{0.763}}& 0.616 &0.626 & 0.691 &0.665 & 0.740 & \textcolor{blue}{\textbf{0.755}}\\
			Success($\uparrow$) & 0.420 & 0.446 &  0.415 & 0.453 & 0.474  & 0.471  &\textcolor{green}{\textbf{0.592 }} & \textcolor{red}{\textbf{0.597}} &0.448  & 0.444  &0.525  &0.504  & 0.579  & \textcolor{blue}{\textbf{0.589}} \\
                        \bottomrule
		\end{tabular}
	}
	\vspace{-3mm}
	\caption{
		\small
		Overall performance on VisEvent dataset \cite{visevent}. Our event-specific model sets a new SOTA record. Our uni-variant, with the same parameter set as in Depth and Thermal, consistently achieves competitive performance across various modalities, leading to a significant margin over the uni-variant of the SOTA modality-specific model \cite{zhu2023visual}.}
	\label{tab-event}
	\vspace{-2mm}
\end{table*}

\noindent \textbf{Comparison on LasHer \cite{lasher}}: Similarly, we conduct evaluations on the LasHer testing set for RGB-T tracker assessment as shown in \cref{tab-lasher}.  Precision (PR) and success rates (SR) are reported following conventional standards \cite{protrack,lasher,zhu2023visual}. Initial comparisons under domain-specific settings reveal the challenges of achieving an architecture-unified model across RGB-D and RGB-T domains, with ProTrack \cite{protrack} and ViPT \cite{zhu2023visual} being the only works consistently leading in both settings. Our Un-Track, following domain-specific finetuning, outperforms the leading ViPT by a significant margin and sets a new SOTA record. In the cross-domain joint learning with a single set of parameters, our uni-model achieves a +3.8\% absolute gain over ViPT. Remarkably, our model with a single set of parameters already achieves very competitive performance compared to the thermal-specific ViPT version.

\begin{figure}[t]
\centering
\includegraphics[width=0.98\linewidth,keepaspectratio]{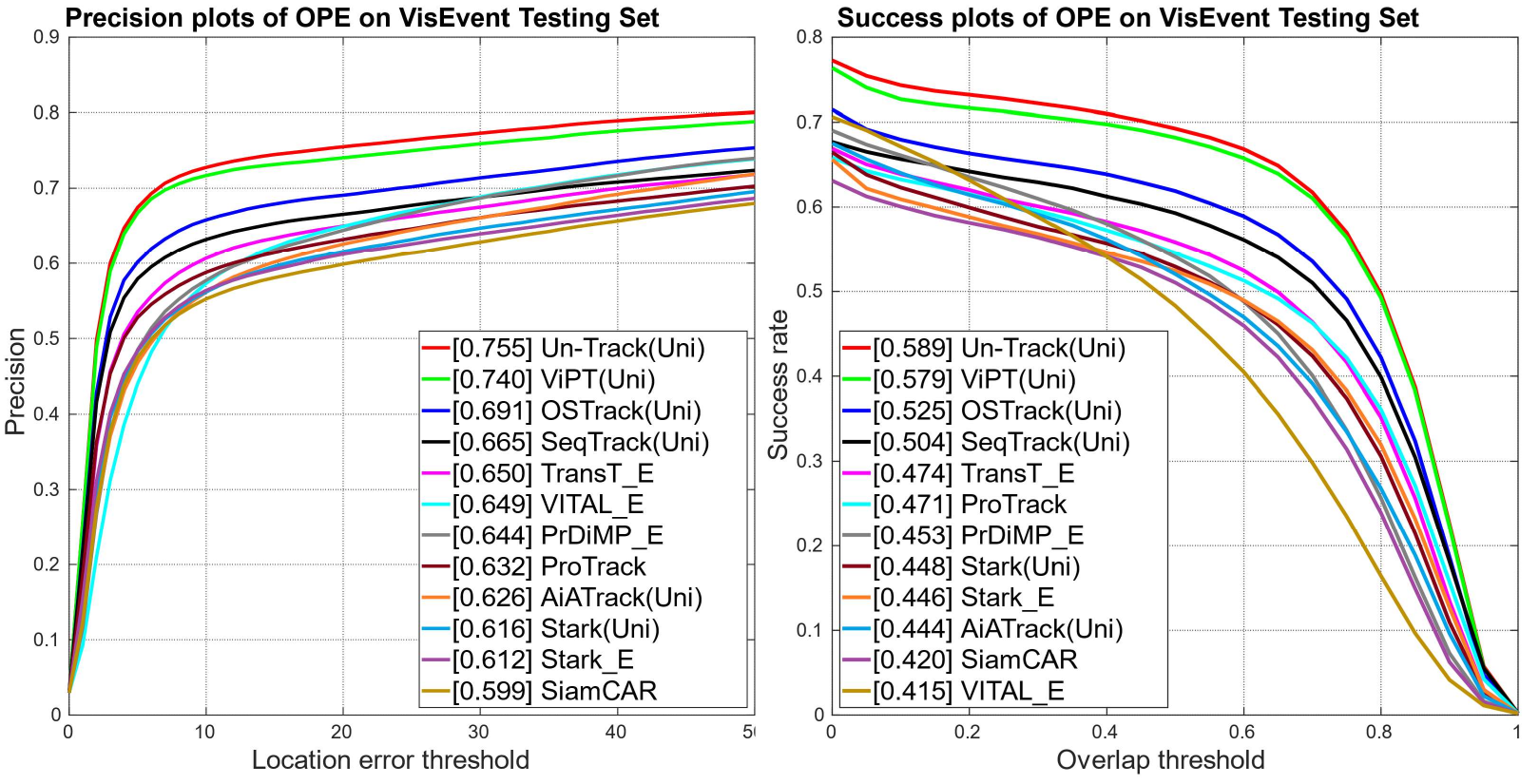}
\vspace{-3.3mm}
\caption{More precision/success comparisons on VisEvent dataset \cite{visevent}. ``Uni" stands for models with a single parameter set. ``\_E" stands for the extension of RGB trackers with event fusion.}\label{fig:visevent}
\vspace{-1mm}
\end{figure}

\noindent \textbf{Comparison on VisEvent \cite{visevent}}: We also evaluate tracker performance with RGB-Event input. Event data, being inherently sparse compared to depth and thermal information, presents challenges in extending existing RGB-D or RGB-T fusion designs for effective integration, hence necessitating specific fusion designs \cite{zhu2023cross,chen20233et,peng2023get}. In contrast, we adopt a unified cross-modal prompting method based on shrinkage fusion. Our approach, with gradual token exchanges between RGB and event modalities, effectively preserves crucial modality-specific clues to enhance feature modeling. Performance-wise, under the event-specific setting, our Un-Track outperforms all counterparts, as shown in \cref{tab-event} and in \cref{fig:visevent}. In the single set of parameters setting, our uni-model achieves a +1.1\% absolute gain in precision over the current SOTA. This underscores the effectiveness of our approach in handling the unique challenges posed by event data integration, which can be mainly contributed to our shared binding that learns the global RGB+X alignment.

\subsection{Generalization Across Datasets} 

In this section, we assess the versatility by evaluating performance on datasets that differ from the training ones, aligning with the goal of achieving a universal model checkpoint applicable to diverse scenarios.

\noindent \textbf{VOT-RGBD2022 \cite{vot22}:} We first perform inference on the VOT-RGBD2022 dataset. Notably, our uni-model achieves superior performance compared to the depth-specific ViPT with a +0.5\% absolute gain in accuracy.

\noindent \textbf{RGBT234 \cite{rgbt234}:}  We also test our model on the other thermal dataset RGBT234, which encompasses sequences with different distributions. Our uni-model surpasses the thermal-specific ViPT with a notable +0.7\% absolute precision gain, as shown in \cref{tab-rgbt234}.

\begin{table*}[t]
	\small
	\centering
	\renewcommand\arraystretch{1.15} 
	\setlength{\tabcolsep}{2.5mm}{ 
		\resizebox{\linewidth}{!}{
			\begin{tabular}{c|cccccccc|ccccc}
                        \toprule
				\small
       & \multicolumn{8}{c|}{Depth-Specific Parameters} & \multicolumn{5}{c}{Uni-model with a Single Set of Parameters}
    \\
                        \midrule
    
				&\tabincell{c}{DRefine\\~\cite{vot21}}&\tabincell{c}{Stark\_D\\~\cite{vot21}}&\tabincell{c}{DMTracker\\~\cite{vot22}}&\tabincell{c}{DeT\\~\cite{depthtrack}}&\tabincell{c}{SBT\_D\\~\cite{vot22}}&\tabincell{c}{SPT\\~\cite{rgbd1k}}&\tabincell{c}{ProTrack\\~\cite{protrack}}&\tabincell{c}{ViPT\\~\cite{zhu2023visual}} &\tabincell{c}{Stark\\ \cite{stark}} &\tabincell{c}{AiATrack\\ \cite{gao2022aiatrack}} &\tabincell{c}{OSTrack\\ \cite{ostrack}}  &\tabincell{c}{SeqTrack \\ \cite{chen2023seqtrack}} &\tabincell{c}{\textbf{Un-Track}\\\textbf{(ours)}}\\
                        \midrule
				EAO($\uparrow$)&0.592&0.647&0.658&0.657&\textbf{\textcolor[rgb]{0,0,1}{0.708}}&0.651&0.651&\textbf{\textcolor[rgb]{1,0,0}{0.721}} &0.445 & 0.641 &0.666 &0.679 & \textbf{\textcolor[rgb]{0,1,0}{0.718}}\\
				Accuracy($\uparrow$)&0.775&0.803&0.758&0.760&\textbf{\textcolor[rgb]{0,0,1}{0.809}}&0.798&0.801&\textbf{\textcolor[rgb]{0,1,0}{0.815}} &0.714 &0.769 &0.808 &0.802 & \textbf{\textcolor[rgb]{1,0,0}{0.820}}\\
				Robustness($\uparrow$)&0.760&0.798&0.851&0.845&\textbf{\textcolor[rgb]{0,0,1}{0.864}}&0.851&0.802&\textbf{\textcolor[rgb]{1,0,0}{0.871}} & 0.598 &0.832 &0.814 &0.846 & \textbf{\textcolor[rgb]{0,1,0}{0.864}}\\
                        \bottomrule
			\end{tabular}
	}}
	\vspace{-3mm}
	\caption{
		\small
		Overall performance on VOT-RGBD2022~\cite{vot22}. Our uni-model, trained on a mix of all modalities, shows robust generalization and outperforms all depth-specific models and other uni-model counterparts.}
	\label{tab-vot22rgbd}
	\vspace{-2mm}
\end{table*}

\begin{table*}[t]
	\small
	\centering
	\renewcommand\arraystretch{1.15} 
	\resizebox{\linewidth}{!}{
		\begin{tabular}{c|cccccccccc| ccccc}
                        \toprule
			\small

             & \multicolumn{10}{c|}{Thermal-Specific Parameters} & \multicolumn{5}{c}{Uni-model with a Single Set of Parameters}
    \\
    \cline{2-16} 
    
			&\tabincell{c}{mfDiMP\\~\cite{mfdimp}} &\tabincell{c}{SGT\\~\cite{sgt}} &\tabincell{c}{DAFNet\\~\cite{dafnet}} &\tabincell{c}{FANet\\~\cite{fanet}}  &\tabincell{c}{MaCNet\\~\cite{macnet}} &\tabincell{c}{CMPP\\~\cite{cmpp}} &\tabincell{c}{APFNet\\~\cite{apfnet}} &\tabincell{c}{ProTrack\\~\cite{protrack}} &\tabincell{c}{ViPT\\~\cite{zhu2023visual}} &\tabincell{c}{\textbf{Un-Track}\\\textbf{(ours)}} 
   
   &\tabincell{c}{Stark\\ \cite{stark}} &\tabincell{c}{AiATrack\\ \cite{gao2022aiatrack}} &\tabincell{c}{OSTrack\\ \cite{ostrack}}  &\tabincell{c}{SeqTrack \\ \cite{chen2023seqtrack}} &\tabincell{c}{\textbf{Un-Track}\\\textbf{(ours)}}\\
			\hline
			MPR($\uparrow$) &0.646 &0.720 &0.796 &0.787 &0.790 &0.823 &0.827 &0.795 &\textcolor{blue}{\textbf{0.835}} &\textcolor{green}{\textbf{0.837}}
   
   & 0.677 &0.711 &0.755 &0.806 & \textbf{\textcolor[rgb]{1,0,0}{0.842}}\\

			MSR($\uparrow$) &0.428 &0.472 &0.544 &0.553 &0.554 &0.575 &0.579 &0.599 &\textcolor{blue}{\textbf{0.617}} & \textcolor{green}{\textbf{0.618}}
   
   & 0.496 &0.508 &0.569 & 0.599 &\textbf{\textcolor[rgb]{1,0,0}{0.625}}\\
                        \bottomrule
		\end{tabular}
	}
	\vspace{-3mm}
	\caption{
		\small
		Overall performance on RGBT234 dataset~\cite{rgbt234}. Our uni-model sets new SOTA records without specific thermal finetuning.}
	\label{tab-rgbt234}
	\vspace{-2mm}
\end{table*}

\begin{figure}[t]
\centering
\includegraphics[width=\linewidth,keepaspectratio]{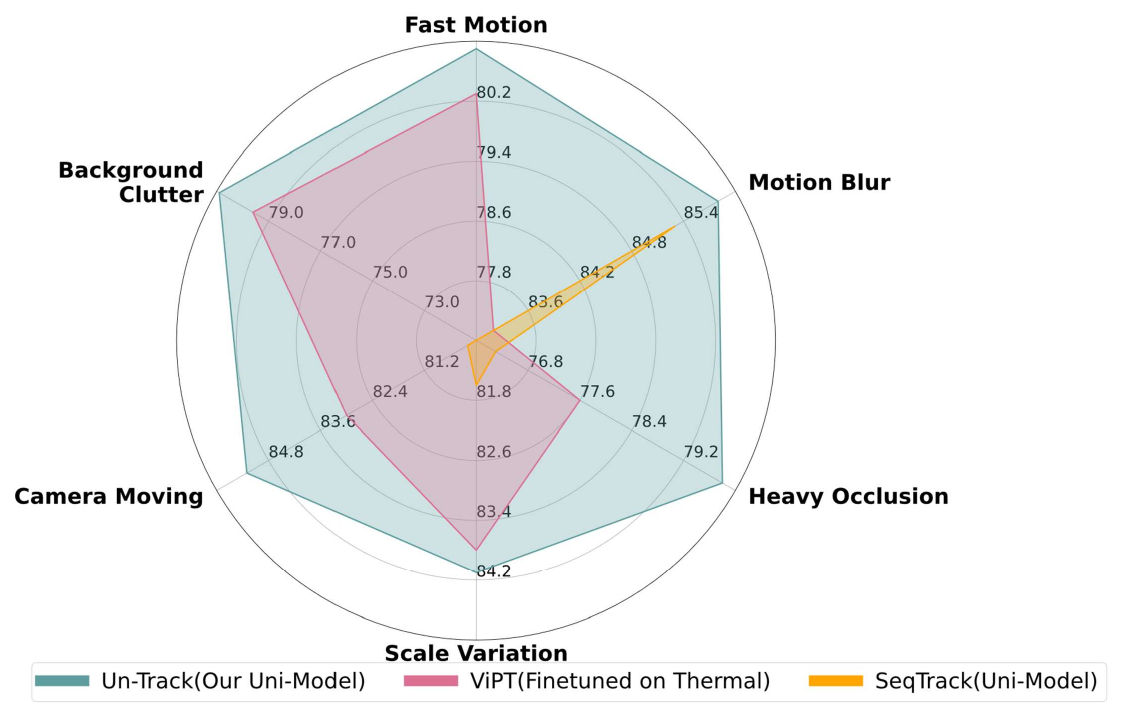}
\vspace{-7mm}
\caption{Per-attribute analysis on the thermal dataset RGBT234 \cite{rgbt234}.  Challenges related to motion and geometry are generally better addressed by event and depth cameras, respectively. Nevertheless, when inferring only with RGB-T data, our Un-Track surpasses both the SOTA thermal-specific method and the current leading uni-tracker. This success underscores our ability to learn emergent alignment across diverse modalities.}
\label{fig:rgbt234}
\vspace{-4mm}
\end{figure}

Moreover, we present a detailed per-attribute comparison with the fine-tuned ViPT in \cref{fig:rgbt234}. We are particularly interested in sequences with motion blur-fast motion-camera moving, as well as sequences with heavy occlusion-scale variants-background clutter. The former motion-related challenges can typically benefit from event clues, renowned for asynchronous computing, while the latter geometry-related challenges can typically benefit from depth cameras. However, these event/depth clues are not directly available in the RGB-T setting. Nevertheless, our Un-Track, trained on all modalities, outperforms both thermal fine-tuned ViPT \cite{zhu2023visual} and the current leading uni-tracker \cite{chen2023seqtrack} with large margins. This underscores our capability in learning event and depth priors through shared binding, without the need for the presence of these modalities during inference.

\noindent \textbf{RGB-only:} In practical scenarios, challenges arise when there are no modal clues available, a typical case is when the auxiliary sensor fails to work properly. We address this demanding case in our study by substituting the modal input with dummy values. Under a such challenging setting, as shown in \cref{tab-rgbonly}, our uni-method consistently outperforms both our RGB baseline fine-tuned counterparts significantly. Notably, such an improvement is achieved with a very limited increase in learning parameters, \ie, +6.65M with +2.14 GFLOPs.

\begin{table}[t]
	\small
	\centering
	\renewcommand\arraystretch{1.15} 
	\setlength{\tabcolsep}{6mm}{
		\resizebox{\linewidth}{!}{
			\begin{tabular}{c|ccc}
                        \toprule
				\small
				&\tabincell{c}{RGB Baseline\\ \cite{ostrack}} &\tabincell{c}{ViPT\\ \cite{zhu2023visual}} &\tabincell{c}{\textbf{Un-Track}\\\textbf{(ours)}} \\
                        \midrule
                        			GFLOPs &21.50&21.80 & 23.64 \\
    			Params (M) &92.08&92.96 & 98.73 \\
       \midrule
				F-score($\uparrow$)&0.529&0.542&0.558 \\
				Re($\uparrow$)&0.522&0.538&0.557 \\
				Pr($\uparrow$)&0.536&0.546&0.560\\

                        \bottomrule
			\end{tabular}
	}}\\
	\vspace{-3mm}
	\caption{
		\small
		Overall performance on DepthTrack test set~\cite{depthtrack} with dummy depth input.}
	\label{tab-rgbonly}
	\vspace{-3mm}
\end{table}

\begin{table}[t]
	\small
	\centering
	\renewcommand\arraystretch{1.15} 
	\setlength{\tabcolsep}{3.7mm}{
		\resizebox{\linewidth}{!}{
			\begin{tabular}{c|ccc}
                        \toprule
				\small
				&\tabincell{c}{w/o\\ Shared Embed} &\tabincell{c}{\cite{zhu2023visual} as \\ Modal Prompt} &\tabincell{c}{w/o\\ LoRA Finetune}\\
                        \midrule
				F-score($\uparrow$)&0.599&0.579&0.594 \\
				Re($\uparrow$)&0.602&0.575&0.598\\
				Pr($\uparrow$)&0.597&0.584&0.596\\
                        \bottomrule
			\end{tabular}
	}}\\
	\vspace{-3mm}
	\caption{
		\small
	Key component analysis.}
	\label{tab-key}
	\vspace{-3mm}
\end{table}

\begin{table*}[t]
\caption{\textbf{Low Rank Approximation} Our plain version is highlighted in \colorbox{Gray}{gray}.}
\label{tab:ablate_lora}
\vspace{-2mm}
\centering
\small
\setlength\tabcolsep{0.5mm}

    \subfloat[
	\label{tab:epoch}
	\textbf{$Rank_k$} (Sec. \ref{sec:recons} )
	]{
		\centering
		\begin{minipage}{0.3\linewidth}{\begin{center}
  \setlength\tabcolsep{6pt}
                    \begin{tabular}{cccc}
                        \toprule
                        & 2 & \colorbox{Gray}{4} & 8 \\ 
                        \midrule
                      F-score($\uparrow$) &  0.607 & 0.610 & 0.602 \\ 
                      Re($\uparrow$) &  0.606 & 0.608 & 0.601 \\ 
                      Pr($\uparrow$) &   0.608 & 0.611 & 0.604 \\ 
                        \bottomrule
                    \end{tabular}
		\end{center}}\end{minipage}
	}
    \subfloat[
	\textbf{$Rank_l$} (Sec. \ref{sec:prompt} )
	\label{tab:bs}
	]{
		\begin{minipage}{0.22\linewidth}{\begin{center}
  \setlength\tabcolsep{6pt}
                    \begin{tabular}{ccc}
                        \toprule
                        4 & \colorbox{Gray}{8} & 16 \\ 
                        \midrule
                        0.596 & 0.610 & 0.606 \\ 
                        0.593 & 0.608 & 0.609 \\ 
                        0.599 & 0.611 & 0.604 \\ 
                        \bottomrule
                    \end{tabular}
		\end{center}}\end{minipage}
	}
    \subfloat[
	\textbf{LoRA} (Sec. \ref{sec:lora} )
	\label{tab:LoRA}
	]{
		\begin{minipage}{0.22\linewidth}{\begin{center}
  \setlength\tabcolsep{8pt}
                    \begin{tabular}{ccc}
                        \toprule
                        2 & \colorbox{Gray}{4} & 8 \\ 
                        \midrule
                        0.601 & 0.610 & 0.600 \\ 
                        0.599 & 0.608 & 0.598\\ 
                        0.602 & 0.611 & 0.602 \\ 
                        \bottomrule
                    \end{tabular}
		\end{center}}\end{minipage}
	}
    \subfloat[
	\textbf{Percentile} (Sec. \ref{sec:prompt} )
	\label{tab:temp}
	]{
		\begin{minipage}{0.22\linewidth}{\begin{center}
    \setlength\tabcolsep{6pt}
                    \begin{tabular}{ccc}
                        \toprule
                        1/8 & \colorbox{Gray}{1/4} & 1/3 \\ 
                        \midrule
                        0.604 & 0.610 & 0.595 \\ 
                        0.606 & 0.608 & 0.593 \\ 
                        0.602 & 0.611 & 0.596 \\ 
                        \bottomrule
                    \end{tabular}
		\end{center}}\end{minipage}
	}
\vspace{-1mm}
\end{table*}

\section{Ablation Studies} 
In this section, we perform all experiments on the DepthTrack testing set \cite{depthtrack} under a single parameter set setting.

\noindent \textbf{Key Component Analysis:} We begin by studying the effectiveness of key components, including the shared embedding, modal prompting, and LoRA finetuning, as summarized in \cref{tab-key}. We initially remove the shared embedding by directly feeding mixed modalities into the learning diagram. It can be seen that this equal treatment of all modalities harms network performance due to the heterogeneous representation across modal domains. Secondly, we replace our prompting block with a recent counterpart that computes fovea attention from additional input \cite{zhu2023visual}. Our designed gradual shrinkage fusion, allowing token-wise interaction, proves to be more effective. We also report performance when we remove the inner fine-tuning with Lora. It can be seen that the performance deteriorates.

\noindent \textbf{Low-Rank:} Low-rank approximation plays a vital role in our model, influencing our shared embedding, modal fusion, and LoRA-finetuning. Hence, the choice of rank is crucial for both the efficiency and effectiveness of our approach. In \cref{tab:ablate_lora}, we systematically explore the impact of low-rank choices within each component.

\noindent \underline{Shared Embedding:} Our objective is to identify an optimal low-rank latent space for merging different modalities effectively. \cref{tab:ablate_lora}(a) presents our ablation study, where we explore ranks of 2, 4, and 8. Lower ranks result in poorer performance due to sparse representations. Conversely, higher ranks capture too many modality-specific details, complicating the search for a shared embedding.

\noindent \underline{Modal Prompting:} As shown in \cref{tab:ablate_lora}(b), similar trends are observed when investigating the low-rank choices for modal prompting with rank values of 4, 8, and 16, as low ranks struggle to capture essential information, while higher ranks introduce an overload of modality-specific details.

\noindent \underline{LoRA-finetuning:} We also vary the ranks between 2, 4, and 8 for the LoRA finetuning technique, as shown in \cref{tab:ablate_lora}(c). Lower ranks exhibited consistently poor performance, while higher ranks in this case tend to result in poorer performance, likely due to overfitting.

\noindent \underline{Remarks:} These experiments emphasize the importance of selecting the best low-rank representations. Nevertheless, our model shows great resilience to the choice of LoRA, showcasing consistent performance across different low-rank configurations. Notably, all our low-rank variants outperform the current SOTA ViPT under the uni-setting, validating our robustness and effectiveness.

\noindent \textbf{Modal Prompting:} During the prompting fusion, a learnable score function is used to categorize tokens into three groups based on their confidence scores. Here, we explore different percentiles for the number of positive, which is the same as the number of negative tokens, leaving the rest as uncertain tokens. As shown in \cref{tab:ablate_lora}(d), the choice of percentile can influence overall performance. Lower percentiles result in poorer performance, as recovering uncertain tokens from very few neighboring tokens is challenging. Higher percentiles also lead to performance degradation since the focus is on token exchange rather than token fusion. The choice of 1/4 leads to the best balance between exchanger and fuser, leading to the best performance.

\noindent \textbf{Shared Embedding:} Here, we perform ablation studies on our shared embedding, a foundational component of our uni-model. The quantitative results are presented in \cref{tab-shared}. We begin by exploring a scenario where our shared embedding is replaced with a variant lacking explicit edge guidance (w/o Explicit Edge). In this configuration, the network learns the shared embedding solely without any edge prior. The results highlight a substantial performance drop, underscoring the pivotal role of explicit edge guidance — a natural and static embedding that binds all modalities together — in facilitating this implicit learning process.

We also conduct experiments using only the explicit edge as the shared embedding, excluding any additional learning modules (w/o Implicit Learning). This approach, too, yields suboptimal performance, primarily due to the neglect of modality-specific clues within each domain.

Finally, we directly compute the low-rank approximation from mixed modalities, bypassing the initial in-domain approximation and subsequent fusion steps (w/o In-domain Approx.). We observe that the network struggles in learning a shared embedding in this direct low-rank approximation setup, primarily due to the intricately mixed representation caused by the domain gap between different modalities.

\begin{table}[t]
	\small
	\centering
	\renewcommand\arraystretch{1.15} 
	\setlength{\tabcolsep}{2mm}{
		\resizebox{\linewidth}{!}{
			\begin{tabular}{c|ccc}
                        \toprule
				\small
				&\tabincell{c}{w/o\\ Explicit Edge} &\tabincell{c}{w/o \\ Implicit Learning} &\tabincell{c}{w/o\\ In-domain Approx.}\\
                        \midrule
				F-score($\uparrow$)&0.600&0.604&0.581 \\
				Re($\uparrow$)&0.602&0.609&0.583\\
				Pr($\uparrow$)&0.598&0.599&0.579\\
                        \bottomrule
			\end{tabular}
	}}\\
	\vspace{-3mm}
	\caption{
		\small
	Ablation on shared embedding}
	\label{tab-shared}
\vspace{-5mm}
\end{table}

\section{Conclusion and Future Work}
We present a successful case of a single-model and any-modality tracker for video object tracking. The proposed method achieves a shared embedding that binds all modalities together, overcoming their heterogeneous representations. This unification is facilitated by lightweight modal prompting and inner finetuning, inheriting benefits from large-scale pre-trained trackers without introducing a substantial computational burden. Exhaustive experiments showcase our improved tracking performance and robust generalization, with any modality input.

\noindent\textbf{Ackowledgement} The authors thank the reviewers and ACs for their tremendous efforts and helpful comments. The event icon is credited to Zuowen Wang. This research is financed in part by the Alexander von Humboldt Foundation, in part by NSFC (62376156, 62322113), and in part by the Ministry of Education and Science of Bulgaria (support for INSAIT, part of the Bulgarian National Roadmap for Research Infrastructure).